\documentclass{llncs}

\usepackage{amsfonts}
\usepackage{amsmath}
\usepackage{arydshln}  
\usepackage[table]{xcolor}  
\usepackage{graphics}  
\usepackage{multirow}  

\usepackage{graphicx}

\usepackage{amssymb}

\usepackage{url}
\usepackage[algo2e,linesnumbered,lined,boxed,vlined]{algorithm2e}

\usepackage{eso-pic} 

\begin{document}

\AddToShipoutPictureBG*{%
  \AtPageUpperLeft{%
    \setlength\unitlength{1in}%
    \hspace*{\dimexpr0.5\paperwidth\relax}
    \makebox(0,-0.75)[c]{\normalsize {\color{black} Accepted for presentation at the 15th International Symposium on Visual Computing (ISVC) 2020, Springer.}}
    }}

\title{Offline versus Online Triplet Mining based on Extreme Distances of Histopathology Patches}
%


\author{Milad Sikaroudi$^\dagger$, Benyamin Ghojogh$^\ddagger$\thanks{The first two authors contributed equally to this work}, Amir Safarpoor$^\dagger$,
Fakhri Karray$^\ddagger$,
Mark Crowley$^\ddagger$,
Hamid R. Tizhoosh$^\dagger$
}

\institute{
$^\dagger$ KIMIA Lab, University of Waterloo, Waterloo, ON, Canada \\
$^\ddagger$Department of Electrical and Computer Engineering,\\ University of Waterloo, Waterloo, ON, Canada  \\
\email{\{msikaroudi, bghojogh, asafarpo, karray, mcrowley, tizhoosh\}@uwaterloo.ca}
}

\maketitle              

\begin{abstract}
We analyze the effect of offline and online triplet mining for colorectal cancer (CRC) histopathology dataset containing 100,000 patches. We consider the extreme, i.e., farthest and nearest patches to a given anchor, both in online and offline mining. While many works focus solely on selecting the triplets online (batch-wise), we also study the effect of extreme distances and neighbor patches before training in an offline fashion. We analyze extreme cases' impacts in terms of embedding distance for offline versus online mining, including easy positive, batch semi-hard, batch hard triplet mining, neighborhood component analysis loss, its proxy version, and distance weighted sampling. We also investigate online approaches based on extreme distance and comprehensively compare offline, and online mining performance based on the data patterns and explain offline mining as a tractable generalization of the online mining with large mini-batch size. As well, we discuss the relations of different colorectal tissue types in terms of extreme distances. We found that offline and online mining approaches have comparable performances for a specific architecture, such as ResNet-18 in this study. Moreover, we found the assorted case, including different extreme distances, is promising, especially in the online approach.  
\keywords{Histopathology, Triplet mining, Extreme distances, Online mining, Offline mining, Triplet network}
\end{abstract}

\section{Introduction}

With the advent of the deep learning methods, image analysis algorithms leveled and, in some cases, surpassed human expert performance. But due to the lack of interpretability, the deep model decision is not transparent enough. Additionally, these models need a massive amount of the labeled data, which can be expensive and time consuming for medical data \cite{tizhoosh2018artificial}. To address the interpretability issue, one may evaluate the performance by enabling consensus, for example, by retrieving similar cases. An embedding framework, such as the triplet loss, can be applied for training models to overcome the expensive label requirement, where either soft or hard similarities can be used \cite{sikaroudi2020supervision}. In triplet loss, triplets of anchor-positive-negative instances are considered where the anchor and positive instances belong to the same class or are similar. Still, the negative instance belongs to another class or is dissimilar to them. Triplet loss aims to decrease and increase the intra-class and inter-class variances of embeddings, respectively, by pulling the anchor and positive closer and pushing the negative away \cite{ghojogh2020fisher}.

Since the introduction of the triplet loss, many updated versions have been proposed to increase efficiency and improve generalization. Furthermore, considering beneficial aspects of these algorithms, such as unsupervised feature learning, data efficiency, and better generalization, the triplet techniques are applied to many other applications, like representation learning in pathology images  \cite{teh2019metric,koch2015siamese,medela2019few,sikaroudi2020supervision} and other medical applications \cite{wang2017multi}. Schroff et al. \cite{schroff2015facenet} proposed a method to encode images into a space with distances reflecting the dissimilarity between instances. They trained a deep neural network using triplets, including similar and dissimilar cases. 


\begin{figure}[!t]
    \makebox[\textwidth][c]{\includegraphics[width=1.4\textwidth]{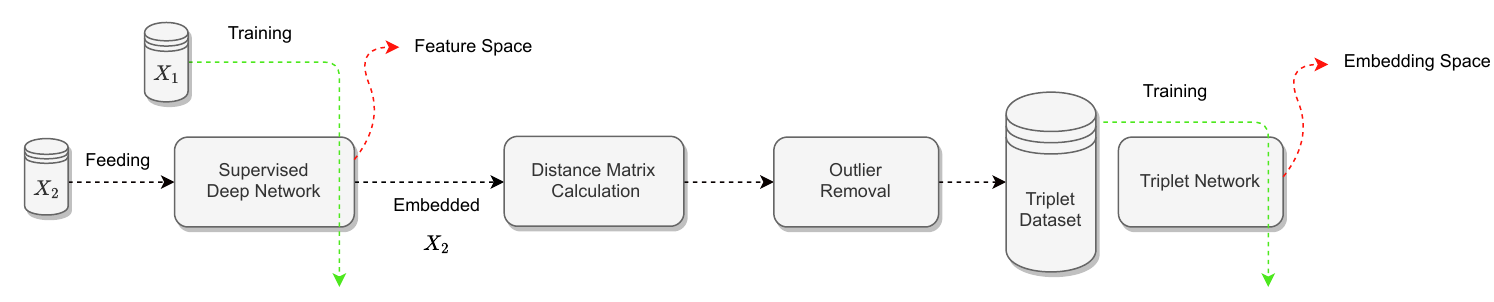}}%
    \caption{Block diagram for the offline triplet mining approach.}
    \label{figure_offline_mining_structure}
\end{figure}

Later, a new family of algorithms emerged to address shortcomings of the triplet loss by selecting more decent triplets with the notion of the similarity for the network while training. These efforts, such as Batch All (BA) \cite{ding2015deep}, Batch Semi-Hard (BSH) \cite{schroff2015facenet}, Batch Hard (BH) \cite{hermans2017defense,peng2019multi}, Neighborhood Components Analysis (NCA) \cite{goldberger2005neighbourhood}, Proxy-NCA (PNCA) \cite{movshovitz2017no,teh2020learning}, Easy Positive (EP) \cite{xuan2020improved}, and Distance Weighted Sampling (DWS) \cite{wu2017sampling} fall into \textit{online triplet mining} category where triplets are created and altered during training within each batch. As the online methods rely on mini-batches of data, they may not reflect the data neighborhood correctly; thus, they can result in a sub-optimal solution. In the \textit{offline triplet mining}, a triplet dataset is created before the training session, while all training samples are taken into account. As a result, in this study, we investigate four offline and five online approaches based on four different extreme cases imposed on the positive and negative samples for triplet generation. 
Our contributions in this work are two-fold. For the first contribution, we have investigated four online methods and the existing approaches, and five offline methods, based on extreme cases. 
Secondly, we will compare different triplet mining methods for histopathology data to analyze them based on their patterns.

The remainder of this paper is organized as follows. 
Section \ref{section_offline_mining} introduces the proposed offline triplet mining methods.
In Section \ref{section_online_mining}, we review the online triplet mining methods and propose new online strategies based on extreme distances. The experiments and comparisons are reported in Section \ref{section_experiments}. Finally, Section \ref{section_conclusion} concludes the paper and reports the possible future work.

\textbf{Notations:} Consider a training dataset $X$ where $x^i$ denotes an instance in the $i$-th class. Let $b$ and $c$ denote the mini-batch size and the number of classes, respectively, and $\mathcal{D}$ be a distance metric function, e.g., squared $\ell_2$ norm. The sample triplet size per class in batch is $w := \lfloor b / c \rfloor$. 
We denote the anchor, positive, and negative instance in the $i$-th class by $x_a^i$, $x_p^i$, and $x_n^i$, respectively, and their deep embeddings by $y_a^i$, $y_p^i$, and $y_n^i$, respectively. 

\section{Offline Triplet Mining}\label{section_offline_mining}

In the offline triplet mining approach, the processing of data is not performed during the triplet network training but beforehand. 
The extreme distances are calculated only once on the whole training dataset, not in the mini-batches. 
The histopathology patterns in the input space cannot be distinguished, especially for the visually similar tissues \cite{jimenez2017analysis}. Hence, we work on the extreme distances in the feature space trained using the class labels. 
The block diagram of the proposed offline triplet mining is depicted in Fig. \ref{figure_offline_mining_structure}. In the following, we explain the steps of mining in detail. 

\textbf{Training Supervised Feature Space:}
We first train a feature space in a supervised manner. For example, a deep network with a cross-entropy loss function can be used for training this space where the embedding of the one-to-last layer is extracted. 
We want the feature space to use the labels to better discriminate classes by increasing their inter-class distances. Hence, We use a set of training data, call it $X_1$, for training the supervised network.

\textbf{Distance Matrix in the Feature Space:}
After training the supervised network, we embed another set of the training data, denoted by $X_2$ (where $X_1 \cup X_2 = X$ and $X_1 \cap X_2 = \varnothing$), in the feature space. 
We compute a distance matrix on the embedded data in the feature space. Therefore, using a distance matrix, we can find cases with extreme distances. We consider every $x \in X_2$ as an anchor in a triplet where its nearest or farthest neighbors from the same and other classes are considered as its positive and negative instances, respectively. We have four different cases with extreme distances, i.e., Easiest Positive and Easiest Negative (EPEN), Easiest Positive and Hardest Negative (EPHN), Hardest Positive, and Easiest Negative (HPEN), and Hardest Positive and Hardest Negative (HPHN). We also have the \textit{assorted} case where one of the extreme cases is randomly selected for a triplet.  

There might exist some outliers in data whose embeddings fall much apart from others. In that case, merely one single outlier may become the hardest negative for all anchors. We prevent this issue by a statistical test \cite{aggarwal2017outlier}, where for every data instance in $X_1$ embedded in the feature space, the distances from other instances are standardized using the $Z$-score normalization. We consider the instances having distances above the $99$-th percentile (i.e., normalized distances above the threshold $2.3263$ according to the standard normal table) as outliers and ignore them. 

\textbf{Training the Triplet Network:}
After preparing the triplets in any extreme case, a triplet network \cite{schroff2015facenet} is trained using the triplets for learning an embedding space for better discrimination of dissimilar instances holding similar instances close enough. We call the spaces learned by the supervised and triplet networks as the feature space and embedding space, respectively (see Fig. \ref{figure_offline_mining_structure}). 

\begin{figure*}[!t]
\centering
\includegraphics[width=5in]{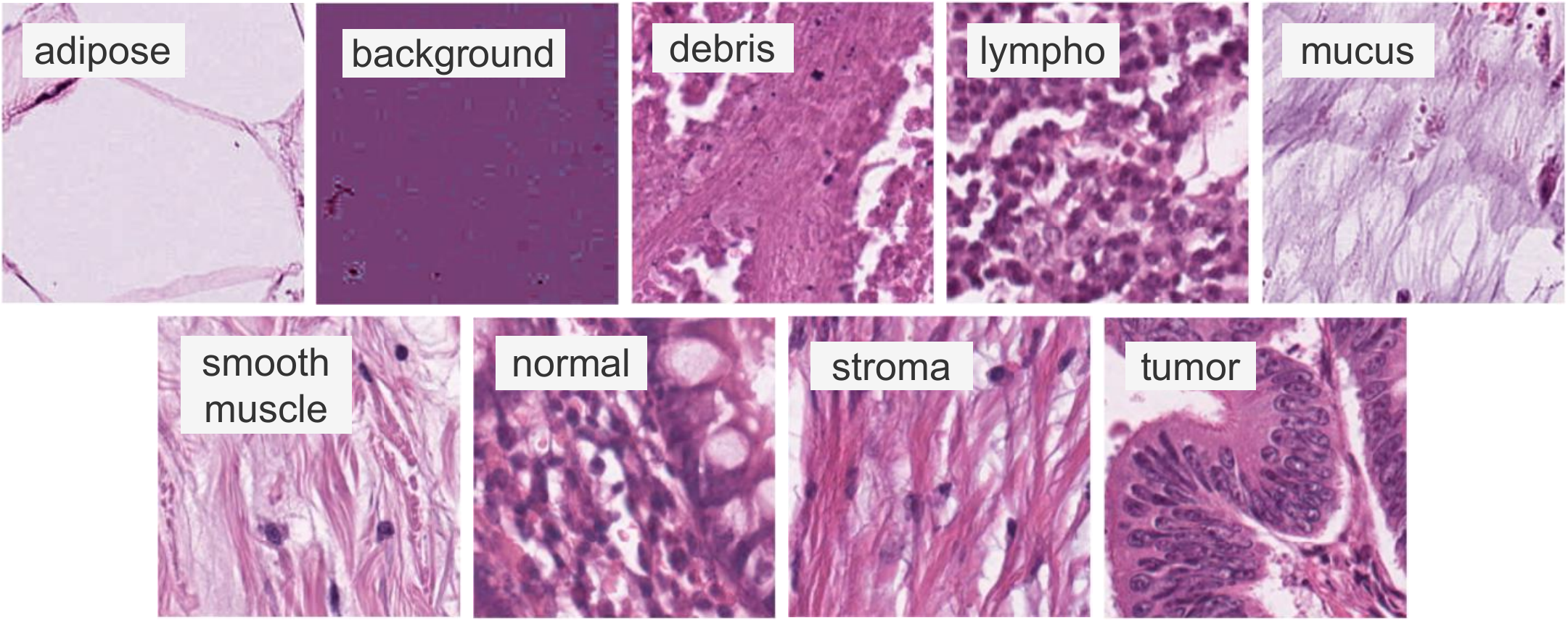}
\caption{Example patches for the different tissue types in the large CRC dataset.}
\label{figure_Examples_of_dataset}
\end{figure*}


\begin{figure}[!t]
    \makebox[\textwidth][c]{\includegraphics[width=1.4\textwidth]{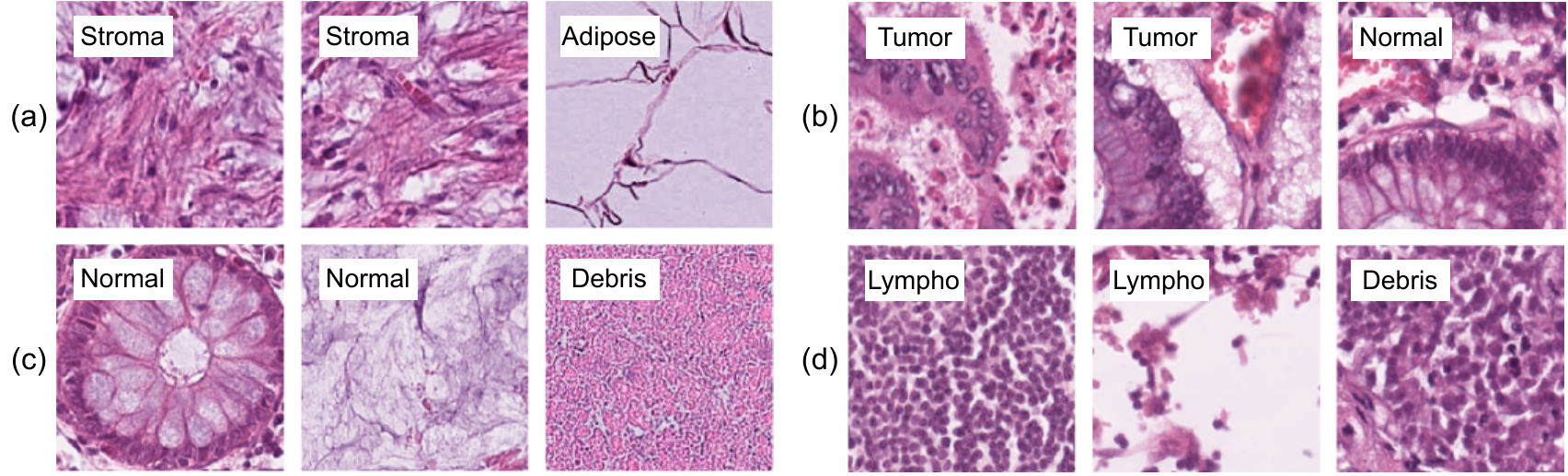}}%
    \caption{Examples for extreme distance triplets: (a) EPEN, (b) EPHN, (c) HPEN, and (d) HPHN.}
\label{figure_extreme_distances}
\end{figure}

\section{Online Triplet Mining}\label{section_online_mining}

In the online triplet mining approach, data processing is performed during the training phase and in the mini-batch of data. 
In other words, the triplets are found in the mini-batch of data and not fed as a triplet-form input to the network. There exist several online mining methods in the literature which are introduced in the following. We also propose several new online mining methods based on extreme distances of data instances in the mini-batch.

\textbf{Batch All \cite{ding2015deep}:} 
One of the online methods which consider all anchor-positive and anchor-negative in the mini-batch. Its loss function is in the regular triplet loss format, summed over all the triplets in the mini-batch, formulated as
\begin{align}
\mathcal{L}_{\text{BA}} := &\sum_{i=1}^{c} \sum_{j=1,\, j \neq i}^{c} \sum_{a=1}^{w} \sum_{p=1,\, p \neq a}^{w} \sum_{n=1}^{w} \!\Big[m + \mathcal{D}(y_{a}^{i}, y_{p}^{i}) - \mathcal{D}(y_{a}^{i}, y_{n}^{j})\Big]_+,
\end{align}
where $m$ is the margin between positives and negatives and $[.]_+ := \max(.,0)$ is the standard Hinge loss. 

\textbf{Batch Semi-Hard \cite{schroff2015facenet}:} 
The hardest (nearest) negative instance in the mini-batch, which is farther than the positive, is selected. Its loss function is 
\begin{align}
\mathcal{L}_{\text{BSH}} &:= \sum_{i=1}^{c}\sum_{a=1}^{w}\sum_{\substack{p=1\\p \neq a}}^{w}\!\Big[m + \mathcal{D}(y_{a}^{i}, y_{p}^{i}) \nonumber \\
&- \underset{j \in \{1,...,c\} \setminus \{i\} \atop {n \in \{1,...,w\}}}{\min} \{\mathcal{D}(y_{a}^{i}, y_{n}^{j}) | \mathcal{D}(y_{a}^{i}, y_{n}^{j})\! > \! \mathcal{D}(y_{a}^{i}, y_{p}^{i}) \}\Big]_+.
\end{align}

\textbf{Batch Hard \cite{hermans2017defense}:} The Hardest Positive and Hardest Negative (HPHN), which are the farthest positive and nearest negative in the mini-batch, are selected. Hence, its loss function is
\begin{align}
\mathcal{L}_{\text{BH}} := \sum_{i=1}^{c}\sum_{a=1}^{w}\!\Big[m&+\underset{p \in \{1,...,w\}\setminus \{a\}}{\max}\mathcal{D}(y_{a}^{i}, y_{p}^{i}) - \underset{j \in \{1,...,c\} \setminus \{i\} \atop {n \in \{1,...,w\}}}{\min}\mathcal{D}(y_{a}^{i}, y_{n}^{j})\Big]_+.
\end{align}

\textbf{NCA \cite{goldberger2005neighbourhood}:} 
The softmax form \cite{ye2019unsupervised} instead of the regular triplet loss \cite{schroff2015facenet} is used. It considers all possible negatives in the mini-batch for an anchor by 
\begin{align}
\mathcal{L}_{\text{NCA}} &:=\! -\! \sum_{i=1}^{c}\sum_{a=1}^{w} \ln\! \Big(\frac{\exp(-\mathcal{D}(y_a^i, y_p^i))}{\sum_{j=1,\, j \neq i}^c \sum_{n=1}^w \exp(-\mathcal{D}(y_a^i, y_n^j))}\Big),
\end{align}
where $\ln(.)$ is the natural logarithm and $\exp(.)$ is the exponential power operator. 


\begin{figure}[!t]
    \makebox[\textwidth][c]{\includegraphics[width=1.2\textwidth]{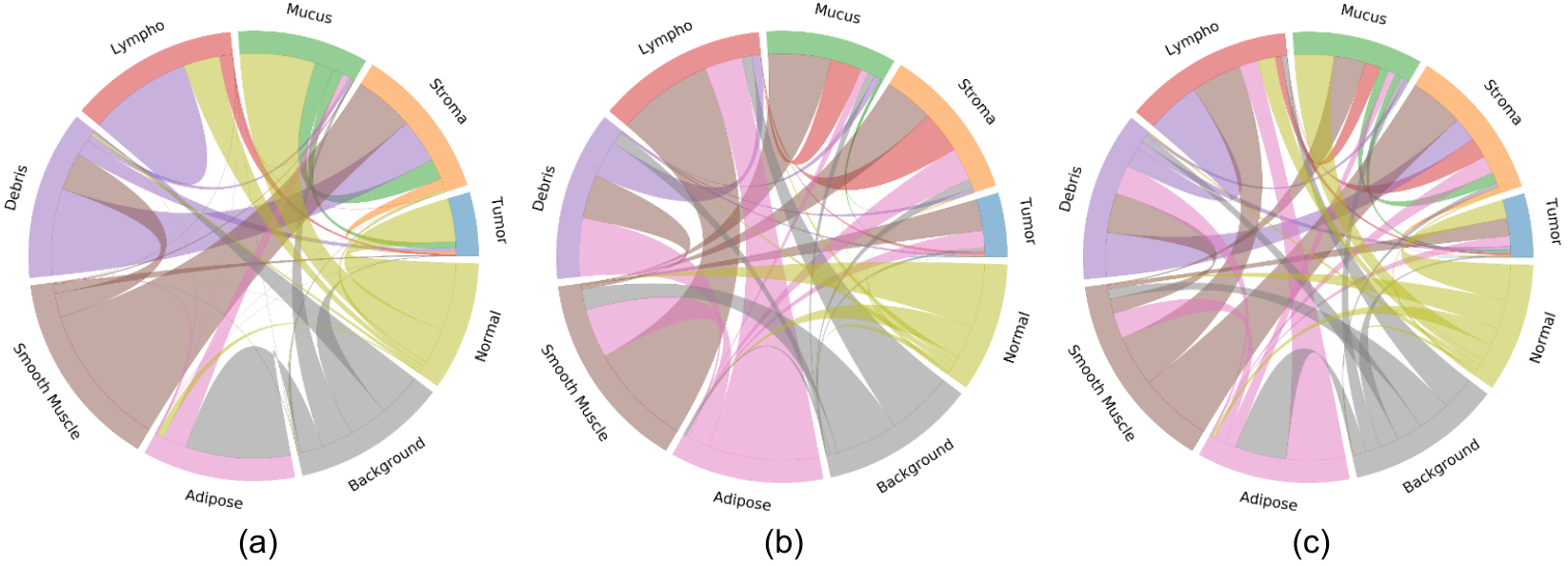}}%
    \caption{Chord diagrams of the negatives in the offline mining based on extreme distances: (a) nearest negatives (EPHN \& HPHN), (b) farthest negatives (EPEN \& HPEN), and (c) \textit{assorted}. The flow from $i$ to $j$ means that $i$ takes $j$ as a negative.}
\label{figure_chords}
\end{figure}

\textbf{Proxy-NCA \cite{movshovitz2017no}:} A set of proxies $\mathcal{P}$, e.g., the center of classes, with the cardinality of the number of classes is used.  An embedding $y$ is assigned to a proxy as $\Pi(y) := \arg \min_{\pi \in \mathcal{P}} \mathcal{D}(y, \pi)$ for memory efficiency. PNCA uses the proxies of positive and negatives in the NCA loss: 
\begin{align}
\mathcal{L}_{\text{PNCA}} := -\! \sum_{i=1}^{c}\sum_{a=1}^{w} \ln\! \Big(\frac{\exp\!\big(\!-\!\mathcal{D}(y_a^i, \Pi(y_p^i))\big)}{\sum_{j=1,\, j \neq i}^c \sum_{n=1}^w \exp\!\big(\!-\!\mathcal{D}(y_a^i, \Pi(y_n^j))\big)}\Big).
\end{align}

\textbf{Easy Positive \cite{xuan2020improved}:} 
Let $y_{ep}^{i} := \arg\min_{p \in \{1,...,w\}\setminus \{a\}} \\ \mathcal{D}(y_{a}^{i}, y_{p}^{i})$ be the easiest (nearest) positive for the anchor. If the embeddings are normalized  and fall on a unit hyper-sphere, the loss in EP method is
\begin{align}\label{equation_loss_EP}
\mathcal{L}_{\text{EP}} := - \sum_{i=1}^{c}\sum_{a=1}^{w} \ln\! \Big(\frac{\exp(y_{a}^{i \top} y_{ep}^{i} )}{\exp(y_{a}^{i \top} y_{ep}^{i} ) + \sum_{j=1,\, j \neq i}^c \sum_{n=1}^w \exp(y_{a}^{i \top} y_{n}^{i} )}\Big).
\end{align}
Our experiments showed that for the colorectal cancer (CRC) histopathology dataset \cite{kather2019predicting}, the performance improves if the inner products in Eq. (\ref{equation_loss_EP}) are replaced with minus distances. We call the EP method with distances by EP-D whose loss function is
\begin{align}\label{equation_loss_EP_D}
&\mathcal{L}_{\text{EP-D}} := - \sum_{i=1}^{c}\sum_{a=1}^{w} \ln\! \Big(\frac{\exp(- \mathcal{D}(y_{a}^{i}, y_{ep}^{i} ))}{\exp(-\mathcal{D}(y_{a}^{i}, y_{ep}^{i}) ) + \sum_{j=1,\, j \neq i}^c \sum_{n=1}^w \exp(-\mathcal{D}(y_{a}^{i}, y_{n}^{i}) )}\Big).
\end{align}

\textbf{Distance Weighted Sampling \cite{wu2017sampling}:} 
The distribution of the pairwise distances is proportional to $q(\mathcal{D}(y_1, y_2)) := \mathcal{D}(y_1, y_2)^{p-2} (1 - 0.25 \mathcal{D}(y_1, y_2)^2)^{(n-3)/2}$ \cite{wu2017sampling}.
For a triplet, the negative sample is drawn as $n^* \sim \mathbb{P}(n | a) \propto \min(\lambda, \\q^{-1}(\mathcal{D}(y_a^i, y_n^j))), \forall j \neq i$. The loss function in the DWS method is
\begin{align}
\mathcal{L}_{\text{DWS}} := \sum_{i=1}^{c}\sum_{a=1}^{w}\sum_{p=1,\, p \neq a}^{w}\!\Big[m + \mathcal{D}(y_{a}^{i}, y_{p}^{i}) - \mathcal{D}(y_{a}^{i}, y_{n^*}) \Big]_+.
\end{align}

\setlength{\tabcolsep}{4pt}
\begin{table*}[!t]
\caption{Results of offline triplet mining on the training and test data}
\label{table_offline}
\centering
\scalebox{0.9}{
\begin{tabular}{l|ccccc|ccccc}
\hline
\multirow{2}{*}{}   & \multicolumn{5}{c}{Train}             & \multicolumn{5}{c}{Test}              \\ \cline{2-11} 
                    & R@1   & R@4   & R@8   & R@16  & Acc.   & R@1   & R@4   & R@8   & R@16  & Acc.   \\ \hline\hline
EPEN    & 92.60 & 97.66 & 98.85 & 99.48 & 95.87 & 89.86 & 96.78 & 98.20 & 99.11 & 94.58 \\ \hline
EPHN    & \textbf{94.82} & \textbf{98.46} & \textbf{99.27} & \textbf{99.70} & \textbf{97.10} & \textbf{94.50} & \textbf{98.41} & \textbf{99.25} & \textbf{99.67} & \textbf{97.21} \\ \hline
HPEN  & 93.22 & 96.93 & 97.71 & 98.35 & 96.16 & 87.11 & 97.01 & 98.83 & 99.59 & 94.10 \\ \hline
HPHN    & 81.62 & 89.73 & 93.15 & 95.78 & 91.19 & 42.71 & 71.07 & 86.13 & 95.32 & 71.25 \\ \hline
\textit{assorted} & 86.40 & 93.65 & 95.93 & 97.66 & 92.53 & 88.56 & 97.31 & 98.95 & 99.52 & 94.60 \\ \hline
\end{tabular}
}
\end{table*}

\setlength{\tabcolsep}{4pt}
\begin{table*}[!t]
\caption{Results of online triplet mining on the training and test data}
\label{table_online}
\centering
\scalebox{0.9}{
\begin{tabular}{l|ccccc|ccccc}
\hline
\multirow{2}{*}{}   & \multicolumn{5}{c}{Train}             & \multicolumn{5}{c}{Test}              \\ \cline{2-11} 
                    & R@1   & R@4   & R@8   & R@16  & Acc.   & R@1   & R@4   & R@8   & R@16  & Acc.   \\ \hline\hline
BA \cite{ding2015deep}                                                               & 95.13 & 98.45 & 99.20 & 99.60                    & 97.73 & 82.42 & 93.94 & 96.93 & 98.58                    & 90.85 \\ \hline
BSH \cite{schroff2015facenet}               & 95.83 & 98.77 & \textbf{99.42} & 99.65                    & 98.00 & 84.70 & 94.78 & 97.34 & 98.75                    & 91.74 \\ \hline
HPHN \cite{hermans2017defense} & 91.52 & 97.14 & 98.60 & 99.34                    & 96.09 & \textbf{86.65} & 95.80 & 97.81 & 99.04                    & 93.20 \\ \hline
NCA \cite{goldberger2005neighbourhood}                                                                     & \textbf{96.45} & \textbf{98.92} & 99.40 & \textbf{99.69}                    & \textbf{98.40} & 78.93 & 92.58 & 96.39 & 98.47                    & 89.65 \\ \hline
PNCA \cite{movshovitz2017no}                                                               & 93.59 & 98.06 & 99.04 & 99.53                    & 97.08 & 80.45 & 93.02 & 96.34 & 98.42                    & 88.72 \\ \hline
EP  \cite{xuan2020improved}                                                            & 84.30 & 94.30 & 96.94 & 98.38 & 92.78 & 74.00 & 90.35 & 95.00 & 97.93 & 85.88 \\ \hline
EP-D                                                              & 86.11 & 95.90 & 97.86 & 99.00                    & 93.30 & 77.23 & 92.14 & 96.18 & 98.49                    & 87.95 \\ \hline
DWS \cite{wu2017sampling}    & 84.43 & 94.78 & 97.27 & 98.59                    & 92.25 & 83.74 & 94.36 & 96.72 & 98.33                    & 92.20 \\ \hline
EPEN                                                         & 87.44 & 95.89 & 97.84 & 98.90                    & 94.03 & 85.48 & 95.40 & 97.65 & 98.92                    & 92.57 \\ \hline
EPHN                                                          & 95.44 & 98.68 & 99.22 & 99.57                    & 97.90 & 85.34 & 94.80 & 97.49 & 98.81                    & 91.77 \\ \hline
HPEN                                                        & 89.53 & 96.67 & 98.21 & 99.15                    & 95.04 & 85.38 & 95.30 & 97.55 & 98.82                    & 92.56 \\ \hline
\textit{assorted}                                                      & 93.73 & 97.98 & 99.02 & 99.57                    & 97.12 & 86.57 & \textbf{96.18} & \textbf{98.25} & \textbf{99.30}                    & \textbf{93.44} \\ \hline
\end{tabular}
}
\end{table*}

\textbf{Extreme Distances:} 
We propose four additional online methods based on extreme distances. We consider every instance once as an anchor in the mini-batch and take its nearest/farthest same-class instance as the easiest/hardest positive and its nearest/farthest other-class instance as the hardest/easiest negative instance. Hence, four different cases, i.e., EPEN, EPHN, HPEN, and HPHN, exist. 
Inspiration for the extreme values, especially the farthest, was the opposition-based learning \cite{tizhoosh2005opposition,tizhoosh2008oppositional}. HPHN is equivalent to BH, which has already been explained. We can also have a mixture of these four cases (i.e., \textit{assorted} case) where for every anchor in the mini-batch, one of the cases is randomly considered. 
The proposed online mining loss functions are as follows:
\begin{align}
\mathcal{L}_{\text{EPEN}} := \sum_{i=1}^{c}\sum_{a=1}^{w}\!\Big[m+\underset{p \in \{1,...,w\}\setminus \{a\}}{\min}\mathcal{D}(y_{a}^{i}, y_{p}^{i})
-\underset{j \in \{1,...,c\} \setminus \{i\} \atop {n \in \{1,...,w\}}}{\max}\mathcal{D}(y_{a}^{i}, y_{n}^{j})\Big]_+, 
\end{align}
\begin{align}
\mathcal{L}_{\text{EPHN}} := \sum_{i=1}^{c}\sum_{a=1}^{w}\!\Big[m+\underset{p \in \{1,...,w\}\setminus \{a\}}{\min}\mathcal{D}(y_{a}^{i}, y_{p}^{i})
-\underset{j \in \{1,...,c\} \setminus \{i\} \atop {n \in \{1,...,w\}}}{\min}\mathcal{D}(y_{a}^{i}, y_{n}^{j})\Big]_+, 
\end{align}
\begin{align}
\mathcal{L}_{\text{HPEN}} := \sum_{i=1}^{c}\sum_{a=1}^{w}\!\Big[m+\underset{p \in \{1,...,w\}\setminus \{a\}}{\max}\mathcal{D}(y_{a}^{i}, y_{p}^{i}) 
-\underset{j \in \{1,...,c\} \setminus \{i\} \atop {n \in \{1,...,w\}}}{\max}\mathcal{D}(y_{a}^{i}, y_{n}^{j})\Big]_+, 
\end{align}
\begin{align}
\mathcal{L}_{\text{Assorted}} := \sum_{i=1}^{c}\sum_{a=1}^{w}\!\Big[m+\underset{p \in \{1,...,w\}\setminus \{a\}}{\min/\max}\mathcal{D}(y_{a}^{i}, y_{p}^{i}) 
-\underset{j \in \{1,...,c\} \setminus \{i\} \atop {n \in \{1,...,w\}}}{\min/\max}\mathcal{D}(y_{a}^{i}, y_{n}^{j})\Big]_+,
\end{align}
where $\min/\max$ denotes random selection between the minimum and maximum operators. 

\section{Experiments and Comparisons}\label{section_experiments}

\textbf{Dataset:}
We used the large colorectal cancer (CRC) histopathology dataset \cite{kather2019predicting} with 100,000 stain-normalized $224\! \times\! 224$ patches. The large CRC dataset includes nine classes of tissues, namely adipose, background, debris, lymphocytes (lymph), mucus, smooth muscle, normal colon mucosa (normal), cancer-associated stroma, and colorectal adenocarcinoma epithelium (tumor).
Some example patches for these tissue types are illustrated in Fig. \ref{figure_Examples_of_dataset}.

\textbf{Experimental Setup:}
We split the data into 70K, 15K, and 15K set of patches, respectively, for $X_1$, $X_2$, and the test data, denoted by $X_t$. We used ResNet-18 \cite{he2016deep} as the backbone of both the supervised network and the triplet network. For the sake of a fair comparison, the mini-batch size in offline and online mining approaches was set to 48 (16 sets of triplets) and 45 ($5$ samples per each of the $9$ classes), respectively, which are roughly equal. The learning rate, the maximum number of epochs, and the margin in triplet loss were $10^{-5}$, $50$, and $0.25$, respectively. The feature-length and embedding spaces were both  $128$.

\begin{figure*}[!t]
\centering
\includegraphics[width=3.8in]{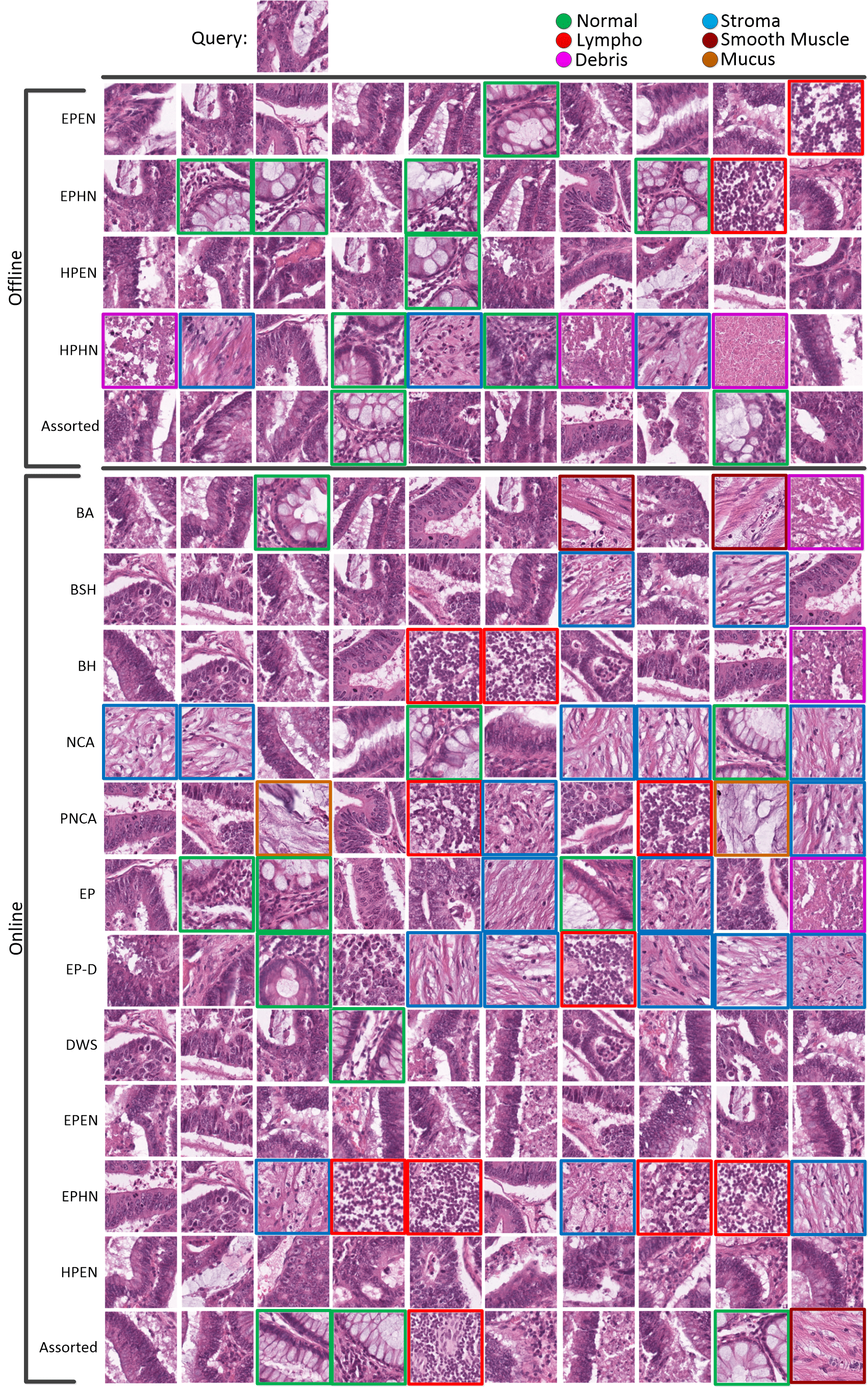}
\caption{The top 10 retrievals (left to right) of a tumor patch query for different loss functions. The patches with no frame are tumor patches.}
\label{figure_query}
\end{figure*}

\textbf{Offline Patches with Extreme Distance:}
Figure \ref{figure_extreme_distances} depicts some examples for the offline created triplets with extreme distances in the feature space. 
The nearest/farthest positives and negatives are visually similar/dissimilar to the anchor patches, as expected. It shows that the learned feature space is a satisfactory subspace for feature extraction, which is reasonably compatible with visual patterns. 

\textbf{Relation of the Colorectal Tissues:}
The chord diagrams of negatives with extreme distances in offline mining are illustrated in Fig. \ref{figure_chords}. 
In both the nearest and farthest negatives, the background and normal tissues have not been negatives of any anchor. Some stroma and debris patches are the nearest negatives for smooth muscle, as well as adipose for background patches, and lymph, mucus, and tumor for normal patches. It stems from the fact that these patches' patterns are hard to discriminate, especially tumor versus normal and stroma and debris versus smooth muscle. In farthest negatives, lymph, debris, mucus, stroma, and tumor are negatives of smooth muscle, as well as debris, smooth muscle, and lymph for adipose texture, and adipose and smooth muscle for normal patches. It is meaningful since they have different patterns. 
Different types of negatives are selected in the \textit{assorted} case, which is a mixture of the nearest and farthest negative patches. It gives more variety to the triplets so that the network sees different cases in training.  

\textbf{Offline versus Online Embedding:}
The evaluation of the embedding spaces found by different offline and online methods are reported in Tables \ref{table_offline} and \ref{table_online}, respectively. The Recall@ (with ranks 1, 4, 8, and 16) and closest neighbor accuracy metrics are reported.

In offline mining, HPHN has the weakest performance on both training and test sets, showing whether the architecture or embedding dimensionality is small for these strictly hard cases or the network might be under-parameterized. We performed another experiment and used ResNet-50  to see whether a more complicated architecture would help \cite{hermans2017defense}. The results showed that for the same maximum number of epochs, either would increase embedding dimensionality to $512$ or utilizing the ResNet-50 architecture increased the accuracy by 4\%. The test accuracy in online mining is not as promising as in offline mining because in online mining we only select a small portion of each class in a mini-batch. The chance of having the most dissimilar/similar patches in a mini-batch is much lower than the case we select triplets in an offline manner. In other words, mining in mini-batches definitely depends upon  a representative population of every class in each batch. 
Besides, the slightly higher training accuracy of the online manner compared to offline mining can be a herald of overfitting in online mining. Tables \ref{table_offline} and \ref{table_online} show that the easiest negatives have comparable results. It is because the histopathology patches (specifically this dataset) may have small intra-class varience for most of the tissue types (e.g., lympho tissue) and large intra-class variance for some others (e.g., normal tissue). Moreover, there is a small inter-class variance in these patches (with similar patterns, e.g., the tumor and normal tissue types are visually similar); hence, using the easy negatives would not drop the performance drastically.  
Moreover, as seen in Fig. \ref{figure_chords}, the hardest negatives might not be perfect candidates for negative patches in histopathology data because many patches from different tissue types erroneously include shared textures in the patching stage \cite{kather2019predicting}. 
In addition to this, the small inter-class variety, explains why the hardest negatives struggle in reaching the best performance, as also reported in \cite{hermans2017defense}. 
Furthermore, literature has shown that the triplets created based on the easiest extreme distances can avoid over-clustering and yield to better performance \cite{xuan2020improved}, which can also be acknowledged by our results. 
The \textit{assorted} approach also has decent performance. Because both the inter-class and intra-class variances are considered. 
Finally, offline and online ways can be compared in terms of batch size. Increasing the batch size can cause the training of the network to be intractable \cite{movshovitz2017no}. On the contrary, a larger batch size implies a better statistical population of data to have a decent representative of every class. An ideal method has a large batch size without sacrificing the tractability. The online approaches can be considered as a special case of offline mining where the mini-batch size is the number of all instances. The offline approach is tractable because of making the triplets in pre-processing. As the Tables \ref{table_offline} and \ref{table_online} show, the offline and online mining approaches have comparable performances. 
The promising performance of online approach has already been investigated by the literature. Here, we also show the promising performance of the offline approach which is because of a good statistical representation for working on the whole data population.
In addition, Table \ref{table_online} shows that the assorted case can result in acceptable embedding because of containing different cases of extreme distances of histopathology patches.


\textbf{Retrieval of Histopathology Patches:}
Finally, in Fig. \ref{figure_query}, we report the top retrievals for a sample tumor query \cite{kalra2020pan}. As the figure shows, EPEN, HPEN, and \textit{assorted} cases have the smallest false retrievals among the offline methods. In online mining, BSH, DWS, EPEN, and HPEN have the best performance. 
These findings coincide with Tables \ref{table_offline} and \ref{table_online} results showing these methods had better  performance. Comparing the offline and online methods in Fig. \ref{figure_query} shows that more number of online approaches than offline ones have false retrievals demonstrating that offline methods benefit from a better statistical population of data. 

\section{Conclusion and Future Direction}\label{section_conclusion}

In this paper, we comprehensively analyzed the offline and online approaches for colorectal histopathology data. We investigated twelve online and five offline mining approaches, including the state-of-the-art triplet mining methods and extreme distance cases. We explained the performance of offline and online mining in terms of histopathology data patterns.
The offline mining was interpreted as a tractable generalization of the online mining where the statistical population of data is better captured for triplet mining. We also explored the relation of the colorectal tissues in terms of extreme distances. 

One possible future direction is to improve upon the existing triplet sampling methods, such as \cite{wu2017sampling}, for online mining and applying that on the histopathology data. One can consider dynamic updates of probabilistic density functions of the mini-batches to sample triplets from the embedding space. This dynamic sampling may improve embedding of histopathology data by exploring more of the embedding space in a stochastic manner.

\bibliographystyle{splncs}      
\bibliography{references.bib}            

\end{document}